\pdfoutput=1

\documentclass[11pt]{article}
\usepackage{naacl22}

\usepackage{times}
\usepackage{latexsym}

\usepackage[T1]{fontenc}
\usepackage[utf8]{inputenc}

\usepackage{microtype}

\usepackage{dsfont}
\usepackage{url}
\usepackage{booktabs} % For formal tables
\usepackage{graphicx}
\usepackage{tabularx}
\usepackage{amsmath}
\usepackage{bbm}
\usepackage{multirow, makecell, caption}
\usepackage{arydshln}
\usepackage{CJKutf8}

\usepackage[normalem]{ulem}
\usepackage{amssymb}
\usepackage{amsfonts}
\usepackage{multicol}

\usepackage{csquotes}
\usepackage{xspace}
\usepackage{paralist}
\usepackage{mdwlist}
\usepackage{subfigure}

\usepackage{enumitem}
\usepackage{cleveref}
\usepackage{bm}
\usepackage{float}

\newcommand{\ie}{\textit{i.e.}}
\newcommand{\eg}{\textit{e.g.}}

\newcommand{\levt}{LevT\xspace}

\newcommand{\editor}{\textsc{EDITOR}\xspace}

\title{\textit{Neighbors Are Not Strangers}: Improving Non-Autoregressive Translation under Low-Frequency Lexical Constraints}

\author{
 Chun Zeng\textsuperscript{\rm $\spadesuit$}\thanks{~~Authors contributed equally.},
 Jiangjie Chen\textsuperscript{\rm $\spadesuit$}\footnotemark[1], 
 Tianyi Zhuang\textsuperscript{\rm $\spadesuit$},
 Rui Xu\textsuperscript{\rm $\spadesuit$}, \\
 \bf Hao Yang\textsuperscript{\rm $\diamondsuit$}, 
 Ying Qin\textsuperscript{\rm $\diamondsuit$}, 
 Shimin Tao\textsuperscript{\rm $\diamondsuit$}, 
 Yanghua Xiao\textsuperscript{\rm $\spadesuit\clubsuit$}\thanks{~~Corresponding author.}\\
\textsuperscript{\rm $\spadesuit$}Shanghai Key Laboratory of Data Science, School of Computer Science, Fudan University\\
\textsuperscript{\rm $\diamondsuit$}Huawei Translation Services Center\\
\textsuperscript{\rm $\clubsuit$}Fudan-Aishu Cognitive Intelligence Joint Research Center\\
\texttt{\{zengc20,jjchen19,tyzhuang20,shawyh\}@fudan.edu.cn},\\
\texttt{ruixu21@m.fudan.edu.cn}, \\
\texttt{\{yanghao30,qinying,taoshimin\}@huawei.com}
}

\begin{document}
\maketitle

\begin{abstract}
Lexically constrained neural machine translation (NMT) draws much industrial attention for its practical usage in specific domains.
However, current autoregressive approaches suffer from high latency.
In this paper, we focus on non-autoregressive translation (NAT) for this problem for its efficiency advantage.
We identify that current constrained NAT models, which are based on iterative editing, do not handle low-frequency constraints well.
To this end, we propose a plug-in algorithm for this line of work, \ie, Aligned Constrained Training (ACT), which alleviates this problem by familiarizing the model with the source-side context of the constraints.
Experiments on the general and domain datasets show that our model improves over the backbone constrained NAT model in constraint preservation and translation quality, especially for rare constraints.\footnote{Our implementation can be found at \url{https://github.com/sted-byte/ACT4NAT}.}
\end{abstract}

\section{Introduction}
\label{sec:intro}
% =======================================

% what's the task? why is it important and challenging?

Despite the success of neural machine translation (NMT) \cite{38ed090f8de94fb3b0b46b86f9133623,NIPS2017_3f5ee243,barrault-etal-2020-findings}, real applications usually require the precise (if not exact) translation of specific terms.
One popular solution is to incorporate dictionaries of pre-defined terminologies as \textit{lexical constraints} to ensure the correct translation of terms, which has been demonstrated to be effective in many areas such as domain adaptation, interactive translation, etc.

% what's the research status quo and unresolved problems

\begin{table}[t]
    \centering
    \small
    \begin{tabular}{|l|}
    \hline
        \textbf{Source} \\
        $
        \underbrace{\mathrm{Travellers}}_\mathrm{1.8K} \underbrace{\mathrm{\textcolor{teal}{screamed}}}_\mathrm{\textcolor{teal}{24}} \underbrace{\mathrm{and}}_\mathrm{2.8M} \underbrace{\mathrm{children}}_\mathrm{30.0K} \underbrace{\mathrm{cried}}_\mathrm{122}.
        $ \\
        \hline
        \textbf{Target} \\
        $
        \underbrace{\mathrm{Reisende}}_\mathrm{944}
        \underbrace{\mathrm{h\ddot{a}tten}}_\mathrm{9.9K}
        \underbrace{\mathrm{\textcolor{brown}{geschrien}}}_\mathrm{\textcolor{brown}{13}}
        \underbrace{\mathrm{und}}_\mathrm{2.6M}
        \underbrace{\mathrm{Kinder}}_\mathrm{20.1K}
        \underbrace{\mathrm{geweint}}_\mathrm{13}.
        $ \\
        % Reisende hätten \textcolor{brown}{geschrien} und Kinder geweint.\\
        \hline
        \textbf{Terminology Constraints} \\
        \textcolor{teal}{scream} $\rightarrow$ \textcolor{brown}{geschrien}\\
        % (\textit{\textcolor{brown}{frequency}: 13})\\
        \hline
        \hline
        \textbf{Unconstrained translation}\\
        Reisende \textit{\textcolor{red}{schrien}} und Kinder rieen. \ \ \ \ \ \ $\Rightarrow$ \textit{{wrong term}}\\
        \hline
        \textbf{Soft constrained translation} \\
        Reisende \textit{\textcolor{red}{rien}}. \ \ $\Rightarrow$ \textit{{incomplete sentence}} \& \textit{{wrong term}}\\
        \hline
        \textbf{Hard constrained translation}\\
        Reisende \textit{\textcolor{brown}{geschrien}}. \ \ \ \ \ \ \ \ \ \ \ \ \ \ \ \ $\Rightarrow$ \textit{{incomplete sentence}} \\
    \hline
    \end{tabular}
    \caption{Translation examples of a lexically constrained non-autoregressive translation (NAT) model \cite{NEURIPS2019_675f9820} under a low-frequency word as constraint.
    The underbraced word frequencies (uncased) are calculated from the vast WMT14 English-German translation (En-De) datasets \cite{NIPS2017_3f5ee243}.
    % ``\textcolor{brown}{geschrien}'' has the frequency of only 13 in a vast corpus in German from WMT14 English$\rightarrow$German translation (En$\rightarrow$De) task \cite{NIPS2017_3f5ee243}.
    % \textcolor{red}{Text in red} means translation error of constraints, and $\Rightarrow$ denotes analysis of the translation errors.
    }
    \label{tab:front}
\end{table}

Previous methods on lexically constrained translation are mainly built upon Autoregressive Translation (AT) models, imposing constraints at inference-time \cite{ture-etal-2012-encouraging,hokamp-liu-2017-lexically,post-vilar-2018-fast} or training-time \cite{luong-etal-2015-addressing,ailem-etal-2021-encouraging}.
However, such methods either are time-consuming in real-time applications or do not ensure the appearance of constraints in the output.
To develop faster MT models for industrial applications, Non-Autoregressive Translation (NAT) has been put forth
% , which is a new paradigm for \textit{parallel} text generation 
\cite{gu2018nonautoregressive,ghazvininejad-etal-2019-mask,NEURIPS2019_675f9820,qian-etal-2021-glancing}, which aims to generate tokens in parallel, boosting inference efficiency compared with left-to-right autoregressive decoding.

% inherent challenges for these problems
% our motivation and idea

Researches on lexically constrained NAT are relatively under-explored.
Recent studies \cite{susanto-etal-2020-lexically,10.1162/tacl_a_00368} impose lexical constraints at inference time upon editing-based iterative NAT models, where constraint tokens are set as the initial sequence for further editing.
However, such methods are vulnerable when encountered with low-frequency words as constraints. 
As illustrated in Table \ref{tab:front}, when translated with a rare constraint, the model is unable to generate the correct context of the term ``geschrien'' as if it does not understand the constraint at all.
It is dangerous since terms in specific domains are usually low-frequency words.
We argue that the main reasons behind this problem are 
\begin{inparaenum}[\it 1)]
    \item the inconsistency between training and constrained inference and 
    \item the unawareness of the source-side context of the constraints.
\end{inparaenum}

To solve this problem, we build our algorithm based on the idea that the context of a rare constraint tends \textit{not} to be rare as well, \ie, ``\textit{a stranger's neighbors are not necessarily strangers}'', as demonstrated in Table \ref{tab:front}.
We believe that, when the constraint is aligned to the source text, the context of its source-side counterpart can be utilized to be translated into the context of the target-side constraint, even if the constraint itself is rare.
Also, when enforced to learn to preserve designated constraints at training-time, a model should be better at coping with constraints during inference-time.

Driven by these motivations, we propose a plug-in algorithm to improve constrained NAT, namely \textbf{A}ligned \textbf{C}onstrained \textbf{T}raining (ACT).
ACT extends the family of editing-based iterative NAT \cite{NEURIPS2019_675f9820,susanto-etal-2020-lexically,10.1162/tacl_a_00368}, the current paradigm of constrained NAT.
Specifically, ACT is composed of two major components: \textit{Constrained Training} and \textit{Alignment Prompting}.
The former extends regular training of iterative NAT with pseudo training-time constraints into the state transition of imitation learning.
The latter incorporates source alignment information of constraints into training and inference, indicating the context of the potentially rare terms.

In summary, this work makes the following contributions:
%\begin{inparaenum}[\it 1)]
\begin{itemize}
    \item We identify and analyse the problems w.r.t. rare lexical constraints in current methods for constrained NAT;
    \item We propose a plug-in algorithm for current constrained NAT models, \ie, aligned constrained training, to improve the translation under rare constraints;
    \item Experiments show that our approach improves the backbone model w.r.t. constraint preservation and translation quality, especially for rare constraints.
\end{itemize}
%\end{inparaenum}

\section{Related Work}
\label{sec:related}

\paragraph{Lexically Constrained Translation}

Existing translation methods impose lexical constraints during either inference or training.
At training time, constrained MT models include code-switching data augmentation \cite{dinu-etal-2019-training, song-etal-2019-code, ijcai2020-496} and training with auxiliary tasks such as token or span-level mask-prediction \cite{ailem-etal-2021-encouraging,lee-etal-2021-improving}.
At inference time, autoregressive constrained decoding algorithms include utilizing placeholder tag \cite{luong-etal-2015-addressing, crego2016systrans}, grid beam search \cite{hokamp-liu-2017-lexically,post-vilar-2018-fast} and alignment-enhanced decoding \cite{alkhouli-etal-2018-alignment, Song_Wang_Yu_Zhang_Huang_Luo_Duan_Zhang_2020, Chen_Chen_Li_2021}. 
% constrained NAT
For the purpose of efficiency, recent studies also focus on non-autoregressive constrained translation.
\citet{susanto-etal-2020-lexically} proposes to modify the inference procedure of Levenshtein Transformer \cite{NEURIPS2019_675f9820} where they disallow the deletion of constraint words during iterative editing.
\citet{10.1162/tacl_a_00368} further develops this idea and introduces a reposition operation that can reorder the constraint tokens.
Our work absorbs the idea of both lines of work.
Based on NAT methods, we brings alignment information by terminologies to help learn the contextual information for lexical constraints, especially the rare ones.

\paragraph{Non-Autoregressive Translation}
% Fully NAT
Although enjoy the speed advantage, NAT models suffer from performance degradation due to the multi-modality problem, \ie, generating text when multiple translations are plausible.
\citet{gu2018nonautoregressive} applies sequence-level knowledge distillation (KD) \cite{kim-rush-2016-sequence} that uses an AT's output as an NAT's new target, which reduces word diversity and reordering complexity in reference, resulting in fewer modes \cite{Zhou2020Understanding,xu-etal-2021-distilled}.
Various algorithms have also been proposed to alleviate this problem, including incorporating latent variables \cite{pmlr-v80-kaiser18a, Shu_Lee_Nakayama_Cho_2020}, 
iterative refinement \cite{ghazvininejad-etal-2019-mask, pmlr-v97-stern19a, NEURIPS2019_675f9820,NEURIPS2020_7a6a74cb}, 
advanced training objective \cite{Wang_Tian_He_Qin_Zhai_Liu_2019, Du2021OAXE} and gradually learning target-side word inter-dependency by curriculum learning \cite{qian-etal-2021-glancing}.
Our work extends the family of editing-based iterative NAT models for its flexibility to impose lexical constraints \cite{susanto-etal-2020-lexically,10.1162/tacl_a_00368}.

\section{Background}
\label{sec:background}

\subsection{Non-Autoregressive Translation}
\label{sec:preliminary}

Given a source sentence as $\bm{x}$ and a target sentence as $\bm{y} = \{y_1, \cdots, y_n\}$, an AT model generates in a left-to-right order, \ie, generating $y_t$ by conditioning on $\bm{x}$ and $y_{<t}$.
An NAT model \cite{gu2018nonautoregressive}, however, discards the word inter-dependency in output tokens, with the conditional independent probability distribution modeled as:
\begin{equation}
\label{eqa:conditional_independency}
    P(\bm{y}|\bm{x})=\prod_{t=1}^{n}P(y_{t}|\bm{x}).
\end{equation}

Such factorization is featured with \textit{high efficiency} at the cost of \textit{performance drop} in translation tasks due to the \textit{multi-modality} problem, \ie, translating in mixed modes and resulting in token repetition, missing, or incoherence.

\subsection{Editing-based Iterative NAT}

Iterative refinement by editing is an NAT paradigm that suits constrained translations due to its flexibility.
It alleviates the multi-modality problem by being autoregressive in editing previously generated sequences while maintaining non-autoregressiveness within each iteration.
Thus, it achieves better performance than fully NATs while is faster than ATs.

\paragraph{Levenshtein Transformer}
\label{subsec:levt}

\begin{table}[t]
    \centering
    \small
    \begin{tabularx}{\linewidth}{|l|X|}
    \hline
        \textbf{Action} & \textbf{Implementation}\\
        \hline
        \multirow{1}*{\textbf{Insertion}} & \textbf{Placeholder Classifier}: predicts the number of tokens ($0 \sim K_{max}$) to be inserted at every consecutive position pairs and then inserts the corresponding number of \texttt{[PLH]}. \\
        & \textbf{Token Classifier}: predicts the actual target token of the \texttt{[PLH]}. \\
        \hline
        \textbf{Deletion} & \textbf{Deletion Classifier}: predicts whether each token (except for the boundaries) should be ``kept'' or ``deleted''. \\
    \hline
    \end{tabularx}
    \caption{The implementation details of insertion and deletion operations in \levt.
    }
    \label{tab:levt_action}
\end{table}

To better illustrate our idea, we use Levenshtein Transformer (\levt, \citealp{NEURIPS2019_675f9820}) as the backbone model in this work, which is a representative model for constrained NAT based on iterative editing.

\levt is based on the Transformer architecture \cite{NIPS2017_3f5ee243}, but more flexible and fast than autoregressive ones.
It models the generation of sentences as Markov Decision Process (MDP) defined by a tuple $(\mathcal{Y}, \mathcal{A}, \mathcal{E}
, \mathcal{R}, {\bm{y}}^0)$.
At each decoding iteration, the agent $\mathcal{E}$ receives an input $ \bm{y} \in \mathcal{Y} $, chooses an action $a \in \mathcal{A}$ and gets reward $r$.
$\mathcal{Y}$ is a set of discrete sentences and $\mathcal{R}$ is the reward function. 
${\bm{y}}^0 \in \mathcal{A}$ is the initial sentence to be edited.

Each iteration consists of two basic operations, \ie, \textit{deletion} and \textit{insertion}, which is described in Table \ref{tab:levt_action}.
For the $k$-th iteration of the sentence $\bm{y}^k = (\texttt{<s>}, y_1, ..., y_n, \texttt{</s>} )$, the insertion consists of placeholder and token classifiers, and the deletion is achieved by a deletion classifier.
\levt trains the model with imitation learning to insert and delete, which lets the agent imitate the behaviors drawn from the expert policy:
\begin{itemize}[noitemsep]
    \item \textbf{Learning to insert}: edit to reference by inserting tokens from a fragmented sentence (\eg, random deletion of reference).
    \item \textbf{Learning to delete}: delete from the insertion result of the current training status to the reference.
\end{itemize}
The key idea is to learn how to edit from a ground truth after adding noise or the output of an adversary policy to the reference.
The ground truth of the editing process is derived from the Levenshtein distance \cite{Levenshtein1965BinaryCC}.

\paragraph{Lexically Constrained Inference}

Lexical constraints can be imposed upon a translation model in:
\begin{inparaenum}[\it 1)]
    \item \textit{soft constraints}: allowing the constraints not to appear in the translation; and
    \item \textit{hard constraints}: forcing the constraints to appear in the translation.
\end{inparaenum}
In NAT, the constraints are generally incorporated at inference time.
\citet{susanto-etal-2020-lexically} injects constraints as the initial sequence for iterative editing in Levenshtein Transformer (\levt, \citealp{NEURIPS2019_675f9820}), achieving soft constrained translation.
And hard constrained translation can be easily done by disallowing the deletion of the constraints.
\citet{10.1162/tacl_a_00368} alters the deletion action in \levt with the reposition operation, allowing the reordering of multiple constraints.

\subsection{Motivating Study: Self-Constrained Translation}
\label{subsec: exploring_the_constraint_frequency}

\begin{figure}[t]
    \centering
    \includegraphics[width=0.9\linewidth]{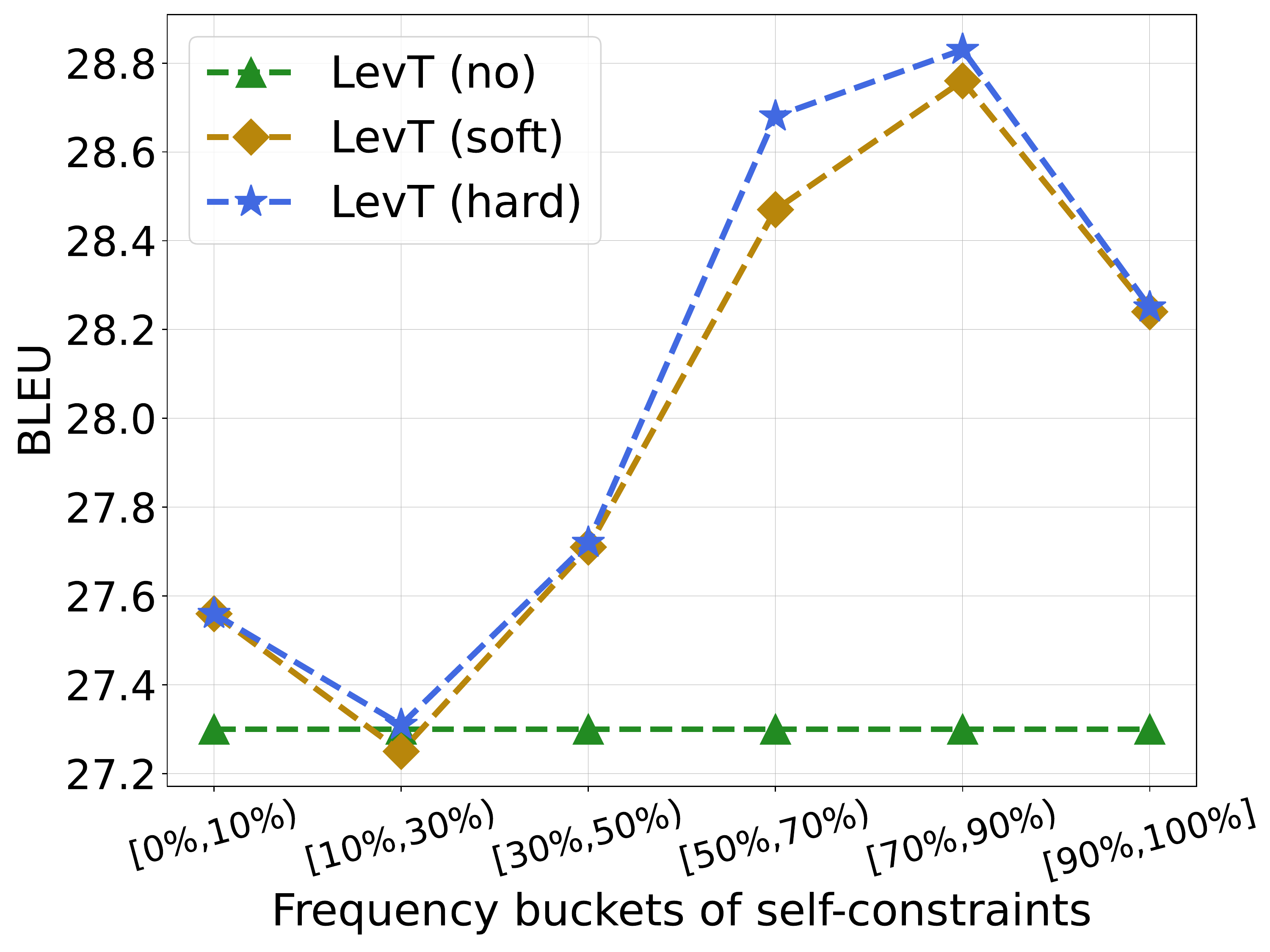}
    \caption{Ablation study of self-constrained translation on WMT14 En$\rightarrow$De test set with Wiktionary terminology constraints \cite{dinu-etal-2019-training}. The absolute average frequencies of self-constraints in a bucket decrease from left to right in the x-axis.}
    \label{fig:self_constr_background}
\end{figure}

According to Table \ref{tab:front}, constrained NAT models seem to suffer from the low-frequency of lexical constraints, which is dangerous as most terms in practice are rare.
To further explore the impact of constraint frequency upon NATs, we conduct a preliminary analysis on constrained \levt \cite{susanto-etal-2020-lexically}.
We sort words in each reference text based on frequency, dividing them into \textit{six} buckets by frequency order (as in Figure \ref{fig:self_constr_background}), and sample a word from each bucket as lexical constraints for translation.\footnote{Sentences that cannot be divided into six buckets are removed.}
We denote these constraints as \textit{self-constraints}.
In this way, we have six times the data, and the six samples derived from one raw sample only differ in the lexical constraints.

As shown in Figure \ref{fig:self_constr_background}, translation performance generally keeps improving as the self-constraint gets rarer.
This is because setting low-frequency words in a sentence as constraints, which are often hard to translate, actually lightens the load of an NAT model.
However, there are two noticeable performance drops around relative frequency ranges of 10\%-30\% (bucket 2) and 90\%-100\% (bucket 6), denoted as \textit{Drop\#1} (-0.3 BLEU) and \textit{Drop\#2} (-0.6 BLEU).
Note that Drop\#1 is mainly due to the the fact that there are mostly unknown tokens (\ie, \texttt{<UNK>}) in the bucket 2.
We leave detailed discussions about buckets and Drop\#1 to Appendix \ref{appendix:bucket}.

In this experiment, we are more interested in the reasons for Drop\#2 when constraints are low-frequency words.
We assume a \textit{trade-off} in \textit{self-constrained} NAT: the model does not have to translate rare words as they are set as an initial sequence (constraints), but it will have a hard time understanding the context of the rare constraint due to 1) the rareness itself and 2) the lack of the alignment information between target-side constraint tokens and source tokens.
Thus, the model does not know how many tokens should be inserted to the left and right of the constraint, which is consistent with the findings in Table \ref{tab:front}.

\section{Proposed Approach}
\label{sec:method}

The findings and assumptions discussed above motivate us to propose a plug-in algorithm for lexically constrained NAT models, \ie, \textbf{A}ligned \textbf{C}onstrained \textbf{T}raining (ACT).
ACT is designed based on two major ideas:
\begin{inparaenum}[\it 1)]
    \item \textit{Constrained Training}: bridging the discrepancy between training and constrained inference;
    \item \textit{Alignment Prompting}: helping the model understand the context of the constraints.
\end{inparaenum}

\subsection{Constrained Training} \label{subsec: CT}

As introduced in $\mathsection$\ref{subsec:levt}, constraints are typically imposed during inference time in NAT \cite{susanto-etal-2020-lexically,10.1162/tacl_a_00368}.
Specifically, lexical constraints are imposed by setting the initial sequence $\bm{y}^0$ as $(\texttt{<S>},  {C}_1, {C}_2, ..., {C}_k, \texttt{</S>})$,
where ${C}_i = (c_1, c_2, ...,c_{l})$ is the $i$-th lexical constrained word, $l$ is the number of tokens in the $i$-th constraint, and $k$ is the number of constraints.

However, such mandatory preservation of the constraints is not carried out during training.
During imitation learning, \textit{random deletion} is applied for ground-truth $\bm{y}^*$ to get the incomplete sentences $\bm{y}'$, producing the data samples for expert policies of how to insert from $\bm{y}'$ to $\bm{y}^*$.
This leads to a situation where the model does not learn to preserve fixed tokens and organize the translation around the tokens.
Such discrepancy could harm the applications of soft constrained translation.

To solve this problem, we propose a simple but effective \textbf{C}onstrained \textbf{T}raining (CT) algorithm.
We first build \textit{pseudo terms} from the target by sampling 0-3 words (more tokens after tokenization) from reference as the pre-defined constraints for training.\footnote{In the experiments, these pseudo constraints are sampled based on TF-IDF score to mimic the rare but important terminology constraints in practice.}
Afterward, we disallow the deletion of pseudo term tokens during building data samples for imitation learning.
This encourages the model to edit incomplete sentences containing lexical constraints into complete ones, bridging the gap between training and inference.

\subsection{Alignment Prompting} 
\label{subsec: ACT}

\begin{figure}[t]
    \centering
    \includegraphics[width=0.9\linewidth]{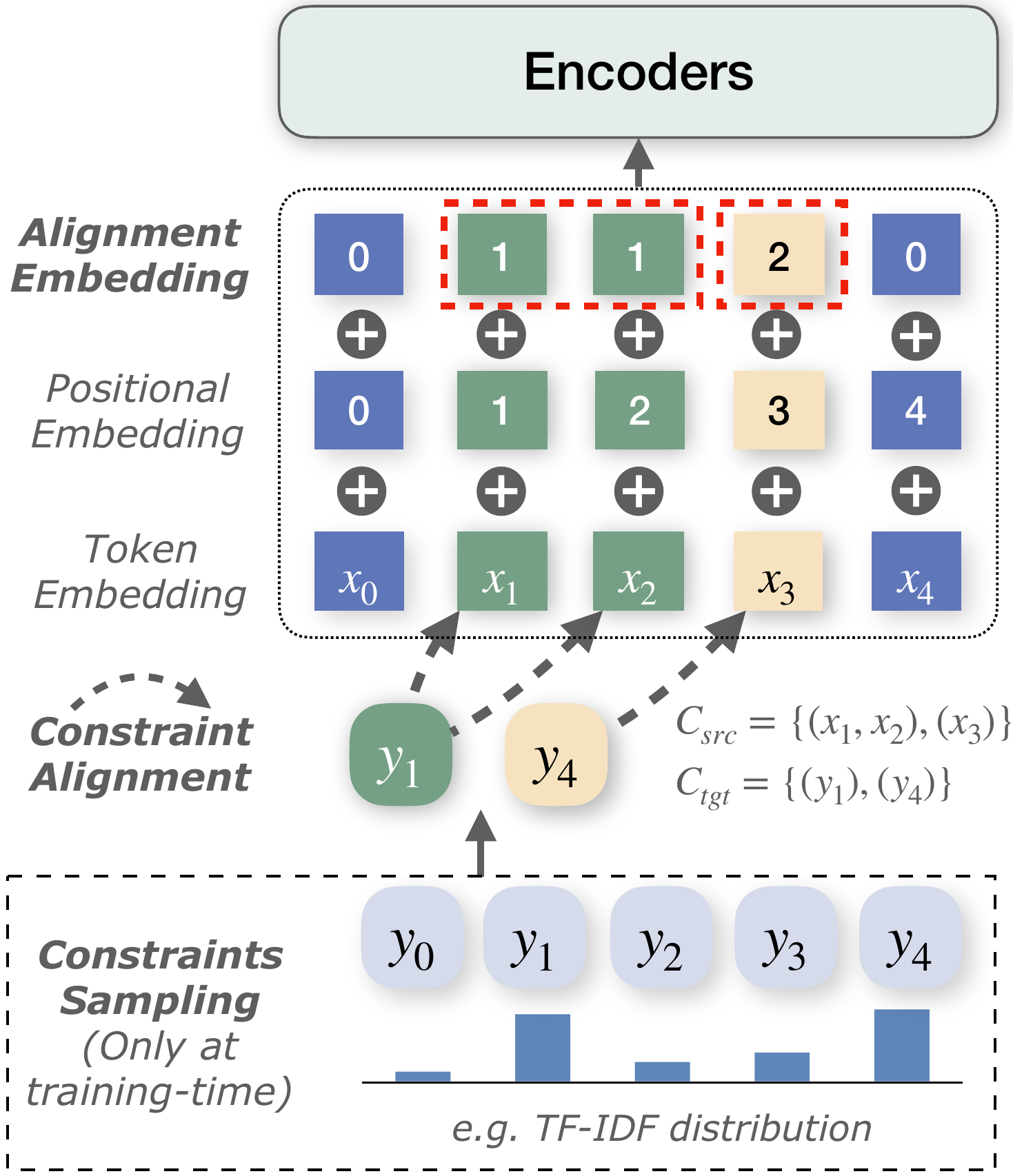}
    \caption{An example of alignment prompting. The constraint tokens $y_\ast$ are given by users during inference, and can also be sampled from target sentence during training.
    Given $y_\ast$, we align them with tokens $x_\ast$ in the source and build alignment embeddings to be fed into the encoder.}
    \label{fig:alignment}
\end{figure}

As stated in $\mathsection$\ref{subsec: exploring_the_constraint_frequency}, we assume the rareness of constraints hinders the model to insert proper tokens of its contexts (\ie, \textit{a stranger's neighbors are also strangers}).
To make the matter worse, previous research \cite{ding2021understanding} has also shown that lexical choice errors on low-frequency words tend to be propagated from the teacher (an AT model) to the student (an NAT model) in knowledge distillation.
% This makes the application of constrained NAT more vulnerable, as the constraints are often low-frequency terms.

However, terminologies, by nature, provide hard alignment information for source and target which the model can conveniently utilize.
Thus, on top of constrained training, we propose an enhanced approach named \textbf{A}ligned \textbf{C}onstrained \textbf{T}raining (ACT).
As illustrated in Figure \ref{fig:alignment}, we propose to directly align the target-side constraints with the source words and prompt the alignment information to the model during both training and inference.

\paragraph{Building Alignment for Constraints}

We first align the source words to the target-side constraints, which are either pseudo constraints during training or actual constraints during inference.
For each translated sentence constraints $\mathcal{C}_\mathrm{tgt}=(C_1,C_2,... ,C_k)$, we use an external alignment tool external aligner, such as GIZA++ \cite{brown-etal-1993-mathematics,och-ney-2003-systematic}, to find the corresponding source words, denoted as $\mathcal{C}_\mathrm{src}=(C'_1, C'_2,... , C'_k)$.

\paragraph{Prompting Alignment into \levt}

The encoder in \levt, besides token embedding and position embedding, is further added with a learnable alignment embedding that comes from $\mathcal{C}_\mathrm{src}$ and $\mathcal{C}_\mathrm{tgt}$.
We set the alignment value for each token in $C'_i$ to $i$ and the others to 0, which are further encoded into embeddings.
The prompting of alignment is not limited to training, as we also add such alignment embeddings to source tokens aligned to target-side constraints during inference.

\section{Experiments}
\label{sec:experiment}

\subsection{Data and Evaluation}

\paragraph{Parallel Data and Knowledge Distillation}

We consider the English$\rightarrow$German (En$\rightarrow$De) translation task and train all of the MT models on WMT14 En-De (3,961K sentence pairs), a benchmark translation dataset.
All sentences are pre-processed via byte-pair encoding (BPE) \cite{sennrich-etal-2016-neural} into sub-word units.
Following the common practice of training an NAT model, we use the sentence-level knowledge distillation data generated by a Transformer, \cite{NIPS2017_3f5ee243} provided by \citet{pmlr-v119-kasai20a}.

\paragraph{Datasets with Lexical Constraints}

\begin{table}[t]
    \centering
    \small
    \begin{tabular}{lccc}
    \toprule
    \textbf{Dataset} & \multirow{2}{*}{\textbf{\# Sent.}} & \textbf{Avg. Len.} & \textbf{Avg. Con.} \\
    (test set) &  & \textbf{of Con.} & \textbf{Freq.} \\
    \midrule
    WMT14-WIKT & 454 & 1.15 & 25,724.73 \\
    % WMT16-WIKT & 270 & 1.11 & 2214.85 \\
    WMT17-IATE & 414 & 1.09 & 3,685.42 \\
    WMT17-WIKT & 728 & 1.22 & 26,252.70 \\
    \hdashline
    OPUS-EMEA  & 2,996 & 1.95 & 2,187.63 \\
    OPUS-JRC  & 2,984 & 1.99 & 3,725.71 \\
    % \hdashline
    % WMT14-Full & 3,003 & - & - \\
    % WMT16-Full & 1,999 & - & - \\
    \bottomrule
    \end{tabular}
    \caption{Statistics of the test sets with target-side lexical constraints.
    ``\textbf{Avg. Len. of Con.}'' denotes the average number of words in a constraint.
    ``\textbf{Avg. Con. Freq.}'' is the average frequency of lexical constraints calculated with the training vocabularies of corresponding language.
    % All test sets are En$\rightarrow$De tasks except for WMT16-WIKT (Ro$\rightarrow$En).
    % Note that WMT14-WIKT are subsets of WMT14,  containing extracted lexical constraints.
    }
\label{tab:datasets}
\end{table}

\begin{table*}[t]
    \small
    \centering
    \begin{tabular}{l||cc|cc|cc|c}
        \hline
        \multirow{2}*{\textbf{Models}} & \multicolumn{2}{c|}{\textbf{WMT17-IATE}} & \multicolumn{2}{c|}{\textbf{WMT17-WIKT}} & 
        \multicolumn{2}{c|}{\textbf{WMT14-WIKT}} & \textbf{Latency} \\
        & \textbf{Term\%} & \textbf{BLEU} & \textbf{Term\%} & \textbf{BLEU} & \textbf{Term\%} & \textbf{BLEU} & (ms)\\ 
        \hline
        \multicolumn{8}{l}{\textit{Reported results in previous work}}\\
        \hline
        Transformer \cite{NIPS2017_3f5ee243}$^\dagger$ & 79.65 & \textbf{29.58} & 79.75 & \textbf{30.80} & \textbf{76.77} & \textbf{31.75} & \textbf{244.5}   \\
        DBA \cite{post-vilar-2018-fast} & 82.00 & 25.30 & \textbf{99.50} & 25.80 & - & - & 434.4  \\
        Train-by-rep \cite{dinu-etal-2019-training} & \textbf{94.50} & 26.00 & 93.40 & 26.30& - & - & -  \\
        \hdashline
        \levt \cite{NEURIPS2019_675f9820}$^\dagger$ & 80.31 & 28.97 & 81.11 & 30.24 & 80.23 & 29.86 & \textbf{92.0}  \\
        \ \ w/ \emph{soft constraint} \cite{susanto-etal-2020-lexically} & 93.81 & 29.73 & 93.44 & 30.82 & 94.43 & 29.93 & -  \\
        \ \ w/ \emph{hard constraint} \cite{susanto-etal-2020-lexically} & \textbf{100.00} & \textbf{30.13} & \textbf{100.00} & \textbf{31.20} & \textbf{100.00} & \textbf{30.49} & -  \\
        \editor \cite{10.1162/tacl_a_00368}$^\dagger$ & 83.00 & 27.90 & 83.50 & 28.80 & - & - & 121.7  \\
        \ \ w/ \emph{soft constraint} & 97.10 & 28.80 & 96.80 & 29.30 & - & - & -  \\ 
        \ \ w/ \emph{hard constraint} & \textbf{100.00} & 28.90 & 99.80 & 29.30 & - & - & 134.1  \\
        \hline 
        \multicolumn{8}{l}{\textit{Our implementation}}\\
        \hline
        LevT$^\dagger$ & 78.32 & \textbf{29.80} & 80.20 & 30.75 & \textbf{79.53} & 29.95 & \textbf{71.9}  \\
        \ \ + constrained training (CT)$^\dagger$ & 78.76 & 29.46 & \textbf{80.77} & \textbf{30.82} & 79.13 & {30.24} & 78.6  \\
        \ \ + aligned constrained training (ACT)$^\dagger$ & \textbf{79.43} & 29.57 & 80.20 & 30.63 & 77.17 & {\textbf{30.35}} & 77.0  \\
        \hdashline
        \levt w/ \emph{soft constraint} & 94.25  & 30.11 & 93.78 & 30.92 & 94.88 & 30.38 & 79.5  \\
        \ \ + constrained training (CT) & 96.24 & 30.19 & 96.61 & 30.96 & 97.44 & {31.01} & \textbf{75.4}  \\
        \ \ + aligned constrained training (ACT) & \textbf{96.90} & {\textbf{30.56}} & \textbf{97.62} & \textbf{31.06} & \textbf{98.82} & {\textbf{31.08}} & 76.3  \\
        \hdashline
        \levt w/ \emph{hard constraint} &\textbf{100.00}& 30.31 &\textbf{100.00}& 30.65 &\textbf{100.00}& 30.49 & 82.7  \\
        \ \ + constrained training (CT) &\textbf{100.00}& 30.31 &\textbf{100.00}& 30.99 &\textbf{100.00}& 31.01 & 78.1  \\
        \ \ + aligned constrained training (ACT) &\textbf{100.00}& \textbf{30.68} &\textbf{100.00}& \textbf{31.18} &\textbf{100.00}& \textbf{31.11} & \textbf{77.0}  \\
        \hline
    \end{tabular}
    \caption{Translation results with lexical constraints.
    \textbf{Term\%} is the constraint term usage rate.
    Method$^\dagger$ translates \textit{without} lexical constraints in input. 
    % {Underlined result} indicates the improvement over \levt is significant ($p < 0.05$).
    }
    \label{tab:main-result}
\end{table*}

Given models trained on the above-mentioned training sets, we evaluate them on the \textit{test sets} of several lexically constrained translation datasets.
These test sets are categorized into two types of standard lexically constrained translation datasets:
\begin{inparaenum}[\it 1)]
    \item Type\#1: tasks from WMT14 \cite{NIPS2017_3f5ee243}
    and WMT17 \cite{bojar-etal-2017-findings}, which are of the same general domain (news) as training sets;
    \item Type\#2: tasks from OPUS \cite{tiedemann-2012-parallel} that are of specific domains (medical and law).
\end{inparaenum}
Particularly, the real application scenarios of lexically constrained MT models are usually domain-specific, and the constrained words in these domain datasets are relatively less frequent and more important.

Following previous work  \cite{dinu-etal-2019-training,susanto-etal-2020-lexically,10.1162/tacl_a_00368}, the lexical constraints in Type\#1 tasks are extracted from existing terminology databases such as Interactive Terminology for Europe (IATE)\footnote{https://iate.europa.eu} and Wiktionary (WIKT)\footnote{https://www.wiktionary.org} accordingly.
The OPUS-EMEA (medical domain) and OPUS-JRC (legal domain) in Type\#2 tasks are datasets from OPUS. 
The constraints are extracted by randomly sampling 1 to 3 words from the reference \cite{post-vilar-2018-fast}.
These constraints are then tokenized with BPE, yielding a larger number of tokens as constraints.
The statistical report is shown in Table \ref{tab:datasets}, indicating the frequencies of Type\#2 datasets are generally much lower than Type\#1 ones.

\paragraph{Evaluation Metrics}

We use BLEU \cite{papineni-etal-2002-bleu} for estimating the general quality of translation. 
We also use \textit{Term Usage Rate} (Term\%,  \citealp{dinu-etal-2019-training,susanto-etal-2020-lexically,lee-etal-2021-improving}) to evaluate lexically constrained translation, which is the ratio of term constraints appearing in the translated text.

\subsection{Models}

We use Levenshtein Transformer (\levt, \citealp{NEURIPS2019_675f9820}) as the backbone model to ACT algorithm for constrained NAT.
We compare our approach with a series of previous MT models on applying lexical constraints:
\begin{itemize}[noitemsep]
    \item \textit{Transformer} \cite{NIPS2017_3f5ee243}, set as the AT baseline;
    \item \textit{Dynamic Beam Allocation} (DBA) \cite{post-vilar-2018-fast} for constrained decoding with dynamic beam allocation over Transformer;
    \item \textit{Train-by-sep} \cite{dinu-etal-2019-training}, trained on augmented code-switched data by replacing the source terms with target constraints or append on source terms during training;
    \item Constrained \levt \cite{susanto-etal-2020-lexically}, which develops \levt \cite{NEURIPS2019_675f9820} by setting constraints as initial editing sequence;
    \item \editor \cite{10.1162/tacl_a_00368}, a variant of \levt, replacing the delete action with a reposition action.
\end{itemize}

\paragraph{Implementation Details}

We use and extend the \texttt{FairSeq} framework \cite{ott-etal-2019-fairseq} for training our models.
We keep mostly the default parameters of \texttt{FairSeq}, such as setting $d_{model}$ = 512, $d_{hidden}$ = 2,048, $n_{heads}$ = 8, $n_{layers}$ = 6 and $p_{dropout}$ = 0.3.
The learning rate is set as 0.0005, the warmup step is set as 4,000 steps. 
All models are trained with a batch size of 16,000 tokens for maximum of 300,000 steps with Adam optimizer \cite{kingma2014method} on 2 NVIDIA GeForce RTX 3090 GPUs with gradient accumulation of 4 batches.
Checkpoints for testing are selected from the average weights of the last 5 checkpoints.
For Transformer \cite{NIPS2017_3f5ee243}, we use the checkpoint released by \citet{ott-etal-2018-scaling}.

\subsection{Main Results}
\label{sec:main_results}

Table~\ref{tab:main-result} reports the performance of \levt with ACT (as well as the CT ablation) on the type 1 tasks (WIKT and IATE as terminologies), compared with baselines.
In general, the results indicate the proposed CT/ACT algorithms achieve a consistent gain in performance, term coverage, and speed over the backbone model mainly in the setting of constrained translation.

When translating with \textit{soft} constraints, \ie, the constraints need not appear in the output, adding ACT to \levt helps preserve the terminology constraints (+$\sim$5 Term\%) and improves translation performance (+0.31-0.88 on BLEU).
If we enforce \textit{hard} constraints, the term usage rate doubtlessly reaches 100\%, with reasonable improvements on BLEU.
When translating \textit{without} constraints, however, adding ACT does not bring consistent improvements as hard and soft constraints do.

% CT
As for the ablation for CT and ACT, we have two observations: 
1) term usage rate increases mainly because of CT, and can be further improved by ACT;
2) translation quality (BLEU) increases due to the additional hard alignment of ACT over CT.
The former could be attributed to the behavior of \textit{not deleting the constraints} in CT.
The latter is because of the introduction of source-side information of constraints that familiarize the model with the constraint context.

Table \ref{tab:main-result} also shows the efficiency advantage of non-autoregressive methods compared with autoregressive ones, which is widely reported in the NAT research literature.
The proposed methods do not cause drops in translation speed against the backbone \levt.
When translating with lexical constraints, \levt with CT or ACT is even faster than \levt.
In contrast, constrained decoding methods for autoregressive models (\ie, DBA) nearly double the translation latency.
Since the main purpose of non-autoregressive research is developing efficient algorithms, such findings could facilitate the industrial usage for constrained translation.

\begin{table}
    \centering
    \small
    \begin{tabular}{lccccc}
        \toprule
        \multirow{2}*{\textbf{Model}} & \multicolumn{2}{c}{\textbf{OPUS-EMEA}} & \multicolumn{2}{c}{\textbf{OPUS-JRC}} \\
        \cmidrule(lr){2-3}
        \cmidrule(lr){4-5}
        & \multicolumn{1}{c}{Term\%} & BLEU & \multicolumn{1}{c}{Term\%} & BLEU \\ 
        \midrule
        % Transformer$^\dagger$ & ? & ? & ? & ? \\
        % \hdashline
        LevT$^\dagger$ & 52.40 & 27.90 & \textbf{55.39} & 30.24 \\
        \ \ + ACT$^\dagger$ & \textbf{53.41} & \textbf{28.30} & 55.35 & \textbf{31.01} \\
        \hdashline
        \levt w/ \emph{soft} & 83.37  & 30.35 & 84.32 & 32.53 \\
        \ \ + ACT & \textbf{92.09} & \textbf{32.02} & \textbf{91.94} & \textbf{33.70} \\
        \hdashline
        \levt w/ \emph{hard} & 100.00 & 30.77 & 100.00 & 30.08 \\
        \ \ + ACT & 100.00 & \textbf{32.30} & 100.00 & \textbf{34.09} \\
        \bottomrule
    \end{tabular}
    \caption{Experiments on test sets from OPUS, which is out of the training domain (WMT14 En$\rightarrow$De).
    Results show that ACT brings larger performance for lower-frequency lexical constraints within these datasets.}
    \label{tab:domain-result}
\end{table}

\paragraph{Translation Results on Domain Datasets}

For a generalized evaluation of our methods, we apply the models trained on the general domain dataset (WMT14 En-De) to medical (OPUS-EMEA) and legal domains (OPUS-JRC).
As seen in Table \ref{tab:domain-result}, even greater performance boosts are witnessed.
When trained with ACT, both term usage (+$\sim$8-10 Term\%) and translation performance (up to 4 BLEU points) largely increase, which is more significant than the general domain.

The reason behind this observation is that the backbone \levt would have a hard time recognizing them as constraints since the lexical constraints in these datasets are much rarer.
Therefore, forcing \levt to translate with these rare constraints would generate worse text, \eg, BLEU drops for 2.45 points on OPUS-JRC than with soft constraints.
And when translating with soft constraints, \levt over-deletes these rare constraints.
In contrast, the context information around constraints is effectively pin-pointed by ACT, so ACT would know the context (``neighbors'') of the rare constraint (``strangers'') and insert the translated context around the lexical constraints.
In this way, more terms are preserved by ACT, and the translation achieves better results.

\section{Analysis}
\label{sec:results}

\begin{figure}[t]
    \centering
	\subfigure[Sorting self-constraints by frequency.] { \label{fig:self_constr_freq} 
	\includegraphics[width=0.9\linewidth]{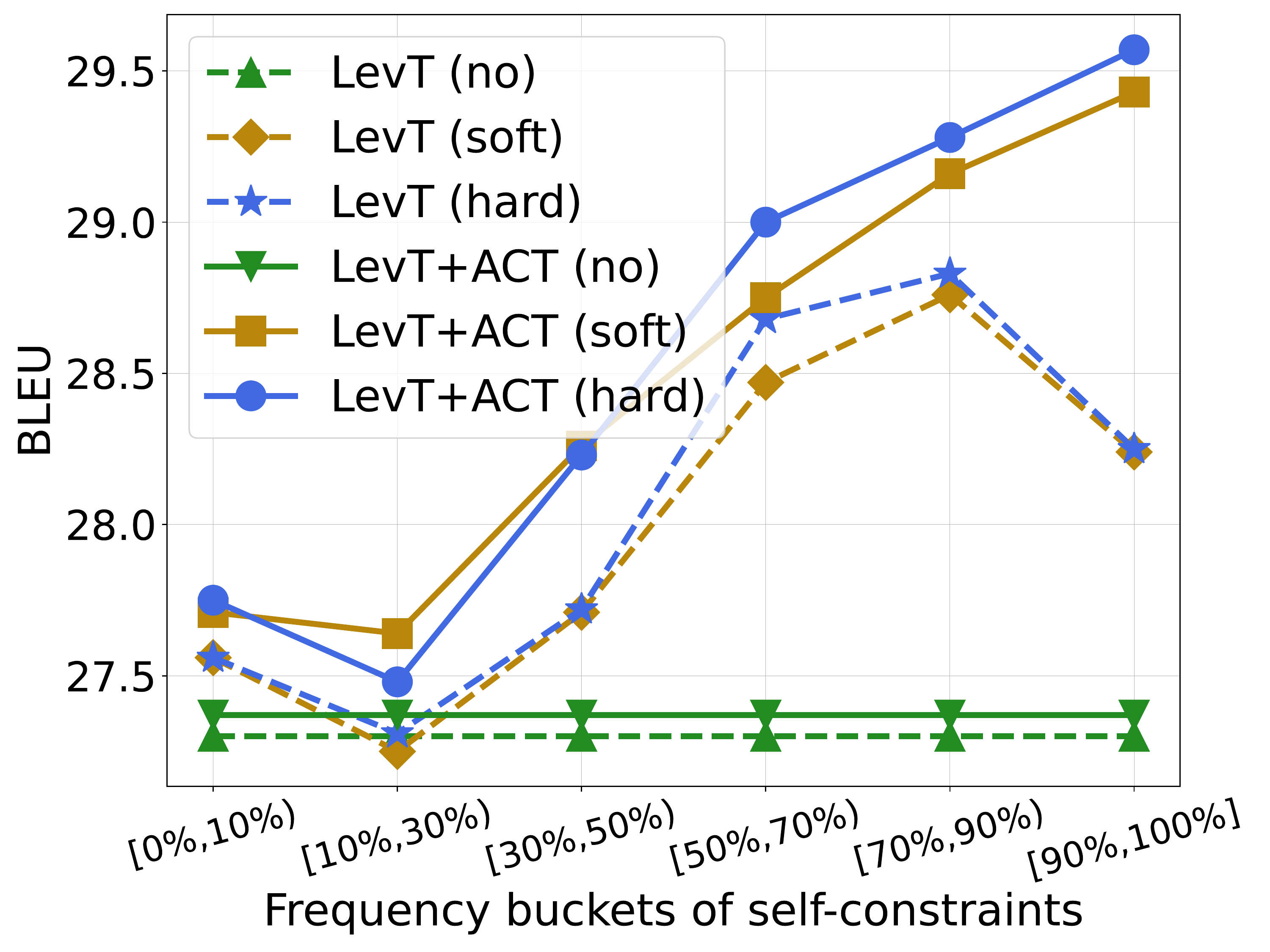}}
	\subfigure[Sorting self-constraints by TF-IDF.] {
	\label{fig:self_constr_tfidf}
	\includegraphics[width=0.9\linewidth]{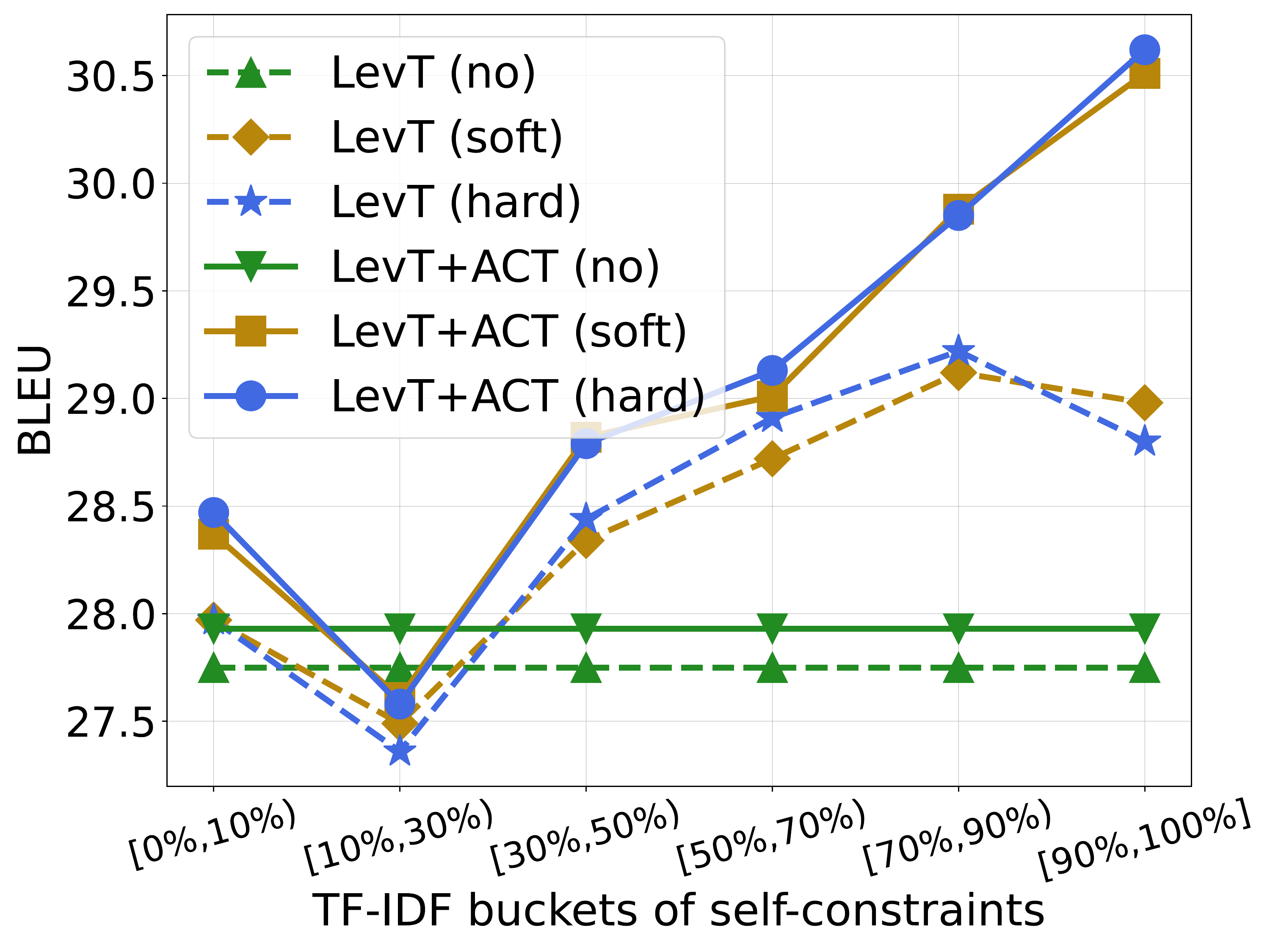}}
    \caption{Extended self-constrained translation results on WMT14-WIKT. Each and every word of a reference is used as a lexical constraint (\ie, self-constraint) for translation, sorted by frequency or TF-IDF.
    }
    \label{fig:self_constr}
\end{figure}

\subsection{Self-Constrained Translation \textit{Revisited}}
\label{sec:self_constr_revisited}

As a direct response to our motivation in this paper, we revisit the ablation study of self-constrained NAT in $\mathsection$\ref{subsec: exploring_the_constraint_frequency} with the proposed ACT algorithm.
Same as before, we build self-constraints from each target sentence and sort them by frequency.
As shown in Figure \ref{fig:self_constr_freq}, different from constrained \levt that suffers from Drop\#2 ($\mathsection$\ref{subsec: exploring_the_constraint_frequency}), ACT managed to handle this scenario pretty well.
Following the motivations given in $\mathsection$\ref{subsec: exploring_the_constraint_frequency}, when constraints become rarer, ACT successfully breaks the \textit{trade-off} with better understanding of the provided contextual information.

\paragraph{What if the self-constraints are sorted based on TF-IDF?}
We also study the importance of different words in a sentence via TF-IDF by forcing them as constraints.
As results in Figure \ref{fig:self_constr_tfidf} show, we have very similar observations from frequency-based self-constraints at Figure \ref{fig:self_constr_freq}, and the gap between \levt and \levt + ACT is even higher as TF-IDF score reaches the highest.

\subsection{How does ACT perform under different kinds of lexical constraints?}

The experiments in $\mathsection$\ref{sec:self_constr_revisited} create pseudo lexical constraints by traversing the target-side reference for understanding the proposed ACT.
In the following analyses, we study different properties of lexical constraints, \eg, frequency and numbers, and how they affect constrained translation.

\begin{figure}[t]
\centering
\includegraphics[width=0.87\linewidth]{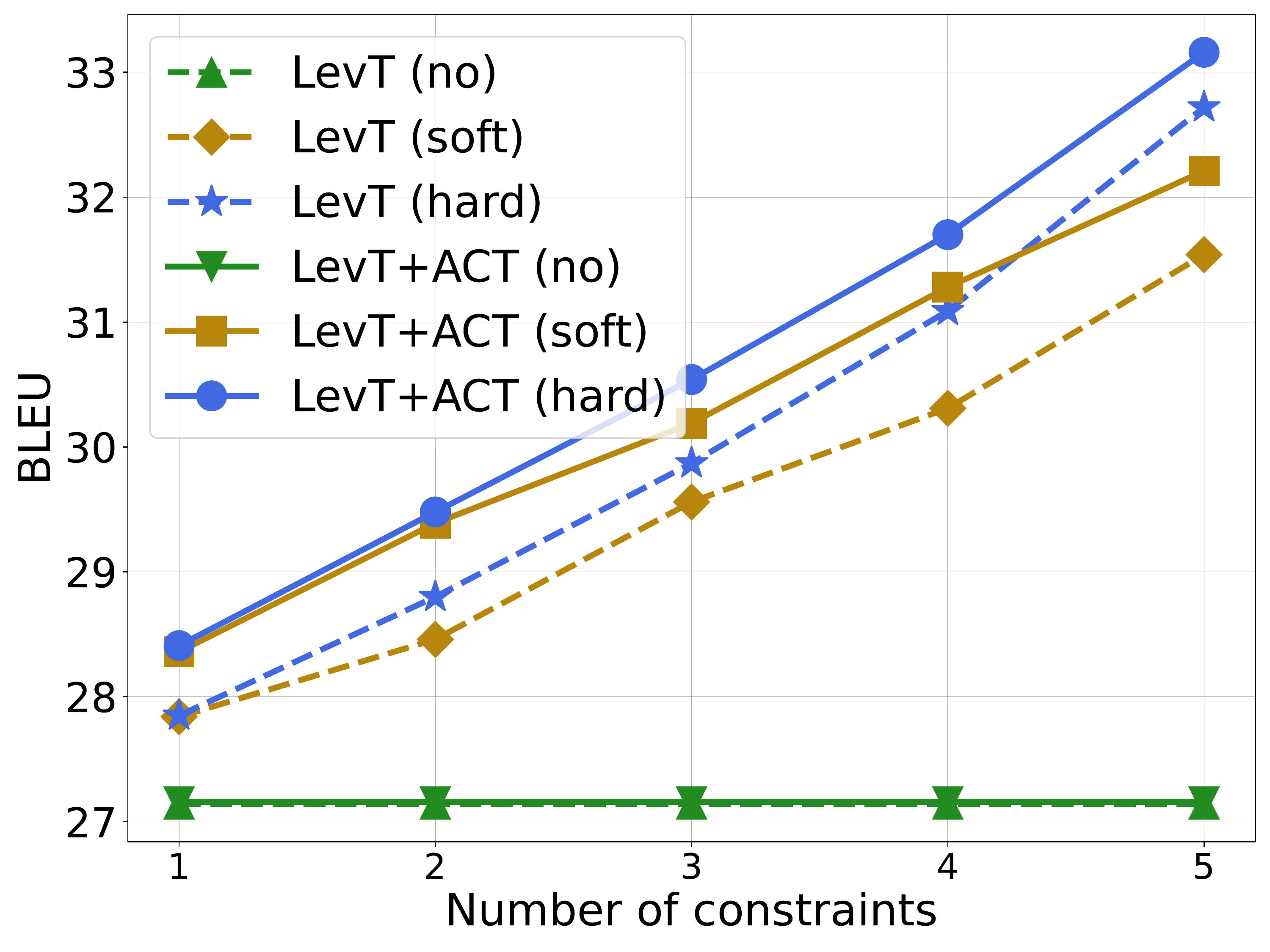}
\caption{Ablation results of constrained translation with one-to-multiple constraints.}
\label{fig:self_constr_number}
\end{figure}

\begin{table*}
    \centering
    \small
    \begin{tabular}{lcccccccccccc}
    \toprule
        \multirow{2}*{\textbf{Model}} & \multicolumn{4}{c}{\textbf{WMT14-WIKT}} & \multicolumn{4}{c}{\textbf{WMT17-IATE}} & \multicolumn{4}{c}{\textbf{WMT17-WIKT}} \\
    \cmidrule(lr){2-5}
    \cmidrule(lr){6-9}
    \cmidrule(lr){10-13}
        & \textsc{All} & \textsc{High} & \textsc{Med.} & \textsc{Low} & 
        \textsc{All} & \textsc{High} & \textsc{Med.} & \textsc{Low} &
        \textsc{All} & \textsc{High} & \textsc{Med.} & \textsc{Low}\\ 
     \midrule
        LevT$^\dagger$ & 29.95 & 30.46 & \textbf{28.03} & 31.49 & \textbf{29.80} &  \textbf{30.08} & \textbf{29.72} & \textbf{29.45} & \textbf{30.75} & \textbf{30.96} & 29.09 & 32.16 \\
        \ \  + ACT$^\dagger$ & \textbf{30.35} & \textbf{30.68} & 28.00 & \textbf{32.54} & 29.57 & 29.63 & 29.57 & 29.20 & 30.63 & 30.35 & \textbf{29.11} & \textbf{32.46} \\
        \hdashline
        \levt w/ \emph{soft}  & 30.38 & 30.37 & 28.50 & 32.19 & 30.11 &  29.25 & 30.67 & 30.04 & 30.92 &  30.70 & \textbf{29.58} & 32.23 \\
        \ \  + ACT  & \textbf{31.08} &  \textbf{30.48} & \textbf{29.18} & {\textbf{33.85}} & \textbf{30.56} &  \textbf{29.93} & \textbf{31.05} & \textbf{30.36} & \textbf{31.06} & \textbf{30.72} & 29.53 & \textbf{32.73} \\
         \hdashline
        \levt w/ \emph{hard}  & 30.49 &  \textbf{30.50} & 28.67 & 31.99 & 30.31 & 29.46 & 30.66 & 30.37 & 30.65 &  30.28 & 29.44 & 32.00 \\
        \ \  + ACT  & \textbf{31.11} & 30.23 & \textbf{29.32} & {\textbf{33.85}} & \textbf{30.68} &  \textbf{29.97} & \textbf{31.18} & \textbf{30.67} & \textbf{31.18} & \textbf{30.58} & \textbf{29.71} & \textbf{32.90} \\
     \bottomrule
    \end{tabular}
    \caption{
    Ablation results of terminology-constrained En$\rightarrow$De translation tasks w.r.t. word frequency of terms.}
    \label{tab:main-result-freq}
\end{table*}

\paragraph{Are improvements by ACT robust against constraints of different frequencies?}

Given terminology constraints in the samples, we sort them by (averaged) frequency and evenly average the corresponding data samples into \textit{high}, \textit{medium} and \textit{low} categories.\footnote{For multi-word terms, we take the average frequency of the words.}
The results on translation quality of each category for the En$\rightarrow$De translation tasks are presented in Table \ref{tab:main-result-freq}.
We find that \levt benef    its mostly from ACT in the scenarios of lower-frequency terms for three datasets.
Although, in some settings such as \textsc{High} in WMT14-WIKT and \textsc{Med} in WMT17-WIKT, the introduction of ACT for constrained \levt seems to bring performance drops for those higher-frequency terms.
Since terms from IATE are rarer than WIKT as in Table \ref{tab:datasets}, the improvements brought by ACT are steady.

\paragraph{Are improvements by ACT robust against constraints of different numbers?}

In more practical settings, the number of constraints is usually more than one.
To simulate this, we randomly sample 1-5 words from each reference as lexical constraints, and results are presented in Figure \ref{fig:self_constr_number}.
We find that, as the number of constraints grows, the translation quality ostensibly becomes better for \levt with or without ACT.
And ACT consistently brings extra improvements, indicating the help by ACT for constrained decoding in constrained NAT.

\subsection{Limitations}

Although the proposed ACT algorithm is effective to improve NAT models on constrained translation, we also find it does not bring much performance gain on translation quality (\ie, BLEU) over the backbone \levt for unconstrained translation.
The results on the full set of WMT14 En$\rightarrow$De test set further corroborate this finding, which is shown in Appendix \ref{appendix:fullwmt}.

Another limitation of our work is that we do not propose a new paradigm for constrained NAT.
The purpose of this work is to enhance existing methods for constrained NAT, \ie, editing-based iterative NAT methods, under rare lexical constraints.
It would be interesting for future research to explore new ways to impose lexical constraints on NAT models, perhaps on non-iterative NAT.

Note that, machine translation in real scenario still falls behind human performance.
Moreover, since we primary focus on improving constrained NAT, real applications calls for refinement in various aspects that we do not consider in this work.

\section{Conclusion}
\label{sec:conclusion}

In this work, we propose a plug-in algorithm (ACT) to improve lexically constrained non-autoregressive translation, especially under low-frequency constraints.
ACT bridges the gap between training and constrained inference and prompts the context information of the constraints to the constrained NAT model. 
Experiments show that ACT improves translation quality and term preservation over the backbone NAT model Levenshtein Transformer.
Further analyses show that the findings are consistent over constraints varied from frequency, TF-IDF, and lengths.
In the future, we will explore the application of this approach to more languages.
We also encourage future research to explore a new paradigm of constrained NAT methods beyond editing-based iterative NAT.

\section*{Acknowledgement}
We would like to thank Xinyao Shen and Shineng Fang at Fudan University as well as Yimeng Chen at Huawei for the support in implementation.
We also thank Rong Ye at ByteDance and the anonymous reviewers for their valuable comments and suggestions for this work.
This work was supported by National Key Research and Development Project (No. 2020AAA0109302), Shanghai Science and Technology Innovation Action Plan (No.19511120400) and Shanghai Municipal Science and Technology Major Project (No.2021SHZDZX0103).

% \clearpage
\bibliography{naacl22}
\bibliographystyle{naacl22}

\clearpage
\appendix

\section{Results on Full Test Set of WMT14 (En$\rightarrow$De)}
\label{appendix:fullwmt}

We extend the experiment on WMT14 En$\rightarrow$De task to the full test set (3,003 samples) in Table~\ref{tab:full}.
Following \citeauthor{susanto-etal-2020-lexically}, we report results on both the filtered test set for sentence pairs that contain at least one target constraint (``Con.'', 454 sentences) and the full test set (``Full'', 3,003 sentences), which contains samples that do not have lexical constraints.
When trained on the full test set, term usage rate raises from 94.88\% to 98.82\% when trained with ACT under soft constrained decoding, but the BLEU score has marginal improvements.
The conclusion is consistent with the experiments in the main body of the paper that \levt with ACT is not significantly better than \levt on unconstrained translation, though our main claim rests on the scenario of constrained NAT.

\begin{table}[h]
    \small 
    \centering
    \begin{tabular}{lccccc}
        \toprule
        \multirow{2}*{\textbf{Model}} & \multirow{2}*{\textbf{Term\%}} & \multicolumn{2}{c}{\textbf{BLEU}} \\
        \cmidrule(lr){3-4}
        & & Full (3,003) & Con. (454) \\
        \midrule
        % Transformer$^\dagger$ & 76.77 & 29.25 & 31.75  \\
        % \hdashline
        LevT$^\dagger$ & \textbf{79.53} & \textbf{26.95} & 29.95  \\
        \ \ + ACT$^\dagger$ & 77.17 & 26.93 & \textbf{30.35}  \\
        \hdashline
        \levt w/ \emph{soft} & 94.88 & 27.04 & 30.38 \\
        \ \ + ACT & \textbf{98.82} & \textbf{27.06} & \textbf{31.08}  \\
        \hdashline
        \levt w/ \emph{hard} & 100.00 & 27.06 & 30.49  \\
        \ \ + ACT & {100.00} & \textbf{27.07} & \textbf{31.11} \\
        \bottomrule
    \end{tabular}
    \caption{Experiments on the test set of WMT14 En$\rightarrow$De task, which shares the same domain of training set. Following \citet{susanto-etal-2020-lexically}, ``Con.'' is the subset of WMT14-Full as shown in Table \ref{tab:datasets}, where every sample has at least one lexical term as constraint.    }
    \label{tab:full}
\end{table}

\begin{table}
    \centering
    \small
    \begin{tabularx}{\linewidth}{|X|}
    \hline
        \multicolumn{1}{|c|}{\textbf{Case 1}}\\
        \hline
        \textbf{Source} \\
        However, carriages are also popular for \textcolor{teal}{hen parties}, he commented.\\
        \hline
        \textbf{Target} \\
        Kutschen sind aber auch für \textcolor{brown}{Jungesellinnenabschiede} beliebt, meint er.\\
        \hline
        \textbf{Terminology Constraints} \\
        \textcolor{teal}{hen parties} $\rightarrow$ \textcolor{brown}{Jungesellinnenabschiede} \\
        \hline
        \hline
        \multicolumn{1}{|c|}{{LevT}}\\
        \hline
        \textbf{Unconstrained translation}\\
        Kutschen sind aber auch für \textit{\textcolor{red}{Hühnerfeiern}} beliebt, kommentierte er.
        \ \ \ \ \ \ \ \ \ \ \ \ \ \ \ \ \ \ \ \ \ \ \ \ \ \ \ \ \ \ \ \ \ \ \ \ \ \ \ \ \ \ \ \ \ \ \ \
        $\Rightarrow$ \textit{{wrong term}}\\
        \hline
        \textbf{Soft constrained translation} \\
        Allerdings sind auch für \textit{\textcolor{red}{Hinnenabschiebeliebt}}, kommentierte er.
        \ \ \ \ \ \ \ \ \ \ \ \ \ \ \ \ \ \ \ \ \ \ \ \ \ \ \ \ \ \ \ \ \ \ \ \ \ \ \ \ \ \ \ \ \ \ \ \ \ \ \ \ \ \ \ 
        $\Rightarrow$  \textit{{wrong term}}\\
        \hline
        \textbf{Hard constrained translation}\\
        Aber Auch für \textit{\textcolor{brown}{Jungesellinnenabschiede}} sind beliebt, sagte er.
        \ \ \ \ \ \ \ \ \ \ \ \ \ \ \ \ \ \ \ \ \ \ \ \ \ \ \ \ \ \ \ \ \ \ \ \ \ \ \ \ \ \ \ \ \ \ \ \ \ \ 
        $\Rightarrow$ \textit{{incomplete sentence}} \\
        \hline
        \hline
        \multicolumn{1}{|c|}{{LevT + ACT}}\\
        \hline
         \textbf{Unconstrained translation}\\
        Wagen sind aber auch für \textit{\textcolor{red}{Hühnerpartys}} beliebt, kommentierte er.
        \ \ \ \ \ \ \ \ \ \ \ \ \ \ \ \ \ \ \ \ \ \ \ \ \ \ \ \ \ \ \ \ \ \ \ \ \ \ \ \ \ \ \ \ \ \ \ \ \ \ \ \ \ \ \
        $\Rightarrow$ \textit{{wrong term}}\\
        \hline
        \textbf{Soft constrained translation} \\
        Kutschen sind aber auch für \textit{\textcolor{brown}{Jungesellinnenabschiede}} beliebt, sagte er.\\
        \hline
        \textbf{Hard constrained translation}\\
        Kutschen sind aber auch für \textit{\textcolor{brown}{Jungesellinnenabschiede}} beliebt, sagte er. \\
        \hline
        \hline
        \hline
        \multicolumn{1}{|c|}{\textbf{Case 2}}\\
        \hline
        \textbf{Source} \\
        The media also reported that several people \textcolor{teal}{injured}.\\
        \hline
        \textbf{Target} \\
        Medien berichteten außerdem von mehreren \textcolor{brown}{Verletzten}.\\
        \hline
        \textbf{Terminology Constraints} \\
        \textcolor{teal}{injured} $\rightarrow$ \textcolor{brown}{Verletzten} \\
        \hline
        \hline
        \multicolumn{1}{|c|}{{LevT}}\\
        \hline
        \textbf{Unconstrained translation}\\
        Die Medien berichteten auch, dass mehrere Menschen \textit{\textcolor{red}{verletzt}} wurden.
        \ \ \ \ \ \ \ \ \ \ \ \ \ \ \ \ \ \ \ \ \ \ \ \ \ \ \ \ \ \ \ \ \ \ \ \ \ \ \ \ \ \ \ 
        $\Rightarrow$ \textit{{wrong term}}\\
        \hline
        \textbf{Soft constrained translation} \\
        Die Medien berichteten auch, dass mehrere \textit{\textcolor{red}{Verletzte}} wurden.\ \ 
        \ \ \ \ \ \ \ \ \ \ \ \ \ \ \ \ \ \ \ \ \ \ \ \ \ \ \ \ \ \ \ \ \ \ \ \ \ \ \ \ \ \ \ \ \ \ \ \ \ \ \ \ \ \ \ \ \ \ \ \ \
        $\Rightarrow$  \textit{{wrong term}}\\
        \hline
        \textbf{Hard constrained translation}\\
        Die Medien berichteten auch, dass mehrere \textit{\textcolor{red}{Verletzte}} wurden. 
        \ \ \ \ \ \ \ \ \ \ \ \ \ \ \ \ \ \ \ \ \ \ \ \ \ \ \ \ \ \ \ \ \ \ \ \ \ \ \ \ \ \ \ \ \ \ \ \ \ \ \ \ \ \ \ \ \ \ \ \ \ \ 
        $\Rightarrow$ \textit{{wrong term}} \\
        \hline
        \hline
        \multicolumn{1}{|c|}{{LevT + ACT}}\\
        \hline
        \textbf{Unconstrained translation}\\
        Die Medien berichteten auch, dass mehrere Menschen \textit{\textcolor{red}{verletzt}} wurden.
        \ \ \ \ \ \ \ \ \ \ \ \ \ \ \ \ \ \ \ \ \ \ \ \ \ \ \ \ \ \ \ \ \ \ \ \ \ \ \ \ \ \ \ \ \  
        $\Rightarrow$ \textit{{wrong term}}\\
        \hline
        \textbf{Soft constrained translation} \\
        Die Medien berichteten auch, dass mehrere \textit{\textcolor{brown}{Verletzten}}.\\
        \hline
        \textbf{Hard constrained translation}\\
        Die Medien berichteten auch, dass mehrere \textit{\textcolor{brown}{Verletzten}}.\\
    \hline
    \end{tabularx}
    \caption{Case study of \levt and \levt with ACT.
    \textcolor{brown}{Text in brown} denotes the constraint word, \textcolor{red}{text in red} denotes the translation error of constraints, and $\Rightarrow$ denotes analysis of the translation errors.
    }
    \label{tab:casestudy}
\end{table}
\section{Case Study}
%In the case of hard constrained translation, \levt is easy to have more interfering words around the constraint words (such as \textit{sind} in case 1).
%In the case of translation with soft or no constraints, \levt has a hard time translating the constraint word accurately.
% After incorporating ACT, constrained training helps iterative editing based on constrained words, and hard alignment of \method better helps the model aware of the context information around low-frequency constrained words.

The case study of \levt and \levt with ACT is presented in Table~\ref{tab:casestudy}.
In the case of unconstrained or soft constrained translation, \levt incorrectly translates low frequency constraint words (\eg, \textit{Hühnerfeiern} in case 1).
In the case of hard constrained translation, \levt tends to have more interfering words around the constraint words (\eg, \textit{sind} in case 1).
% This is because after adding low-frequency constrained words, \levt does not translate well as a whole due to the inability to aware of the semantics of constrained words (\eg, not knowing how many \emph{[PLH]}s should be inserted on the left and right).
After incorporating ACT, we witness consistent improvements in the translation of the constraints for \levt, especially for soft constrained translation where it successfully translates given constraints.
However, when the translation is not constrained on lexical terms (\ie, unconstrained translation), \levt with ACT still struggles at translating the term correctly (both case 1 and 2).

\begin{table*}[t]
    \centering
    %\small
    \begin{tabular}{ccccccc}
        \toprule
        \textbf{Bucket} & \textbf{\# PUNC} & \textbf{\# NN*} & \textbf{\# (JJ*,RB*,VB*)} & \textbf{\# UNK} &\textbf{ \# OTHER} & \textbf{\# ALL} \\ 
        \midrule
        1 & 1,300 & 971 & 433 & 0 & 63 & 2,767 \\
        2 & 148 & 1,520 & 567 & 186 & 346 & 2,767 \\
        3 & 12 & 1,926 & 531 & 97 & 201 & 2,767 \\
        4 & 2 & 2,298 & 308 & 4 & 155 & 2,767 \\
        5 & 0 & 2,377 & 208 & 3 & 179 & 2,767 \\
        6 & 0 & 2,336 & 134 & 5 & 292 & 2,767 \\
        \bottomrule
    \end{tabular}
    \caption{
    The counted statistics of constraint tokens within each bucket in self-constrained translation study, where tokens are categorized according to their Part-Of-Speech tags.
    Among them, 
    PUNC denotes punctuation;
    NN* denotes all sets of nouns (whose POS tags start with NN, including NN, NNP, NNS, NNPS, etc.);
    JJ*, RB* and VB* denotes all kinds of adjectives, adverbs and verbs; 
    UNK is the constraint with \texttt{UNK} token and some special symbols;
    and others as denoted as OTHER.}
    \label{tab:pos-result}
\end{table*}

\begin{figure*}[h]
    \centering
	\subfigure[Sorting self-constraints by frequency.] {
	\label{fig:revision_self_constr_freq} 
 	\includegraphics[width=0.45\linewidth]{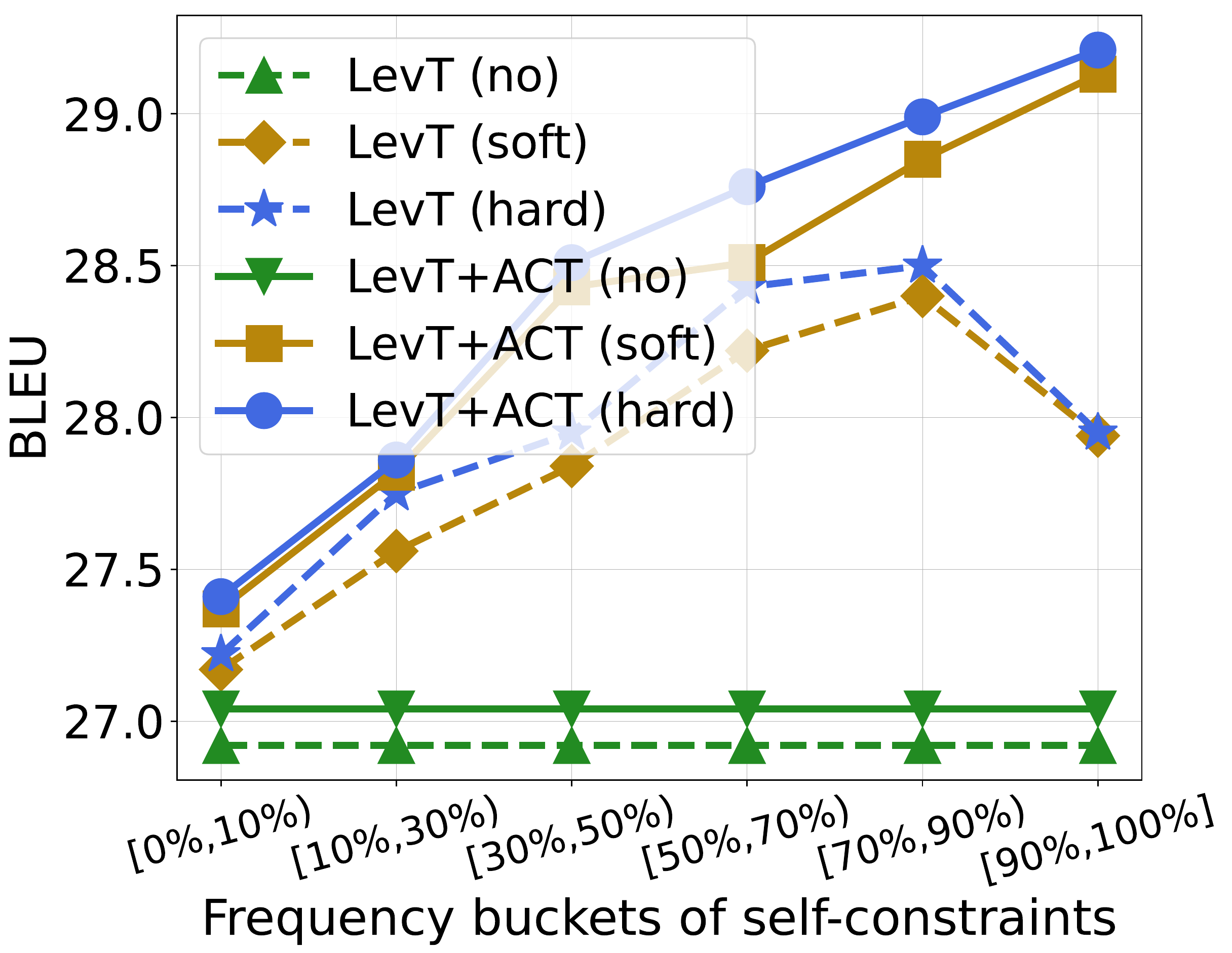}}
	\subfigure[Sorting self-constraints by TF-IDF.] {
	\label{fig:revision_self_constr_tfidf}
	\includegraphics[width=0.45\linewidth]{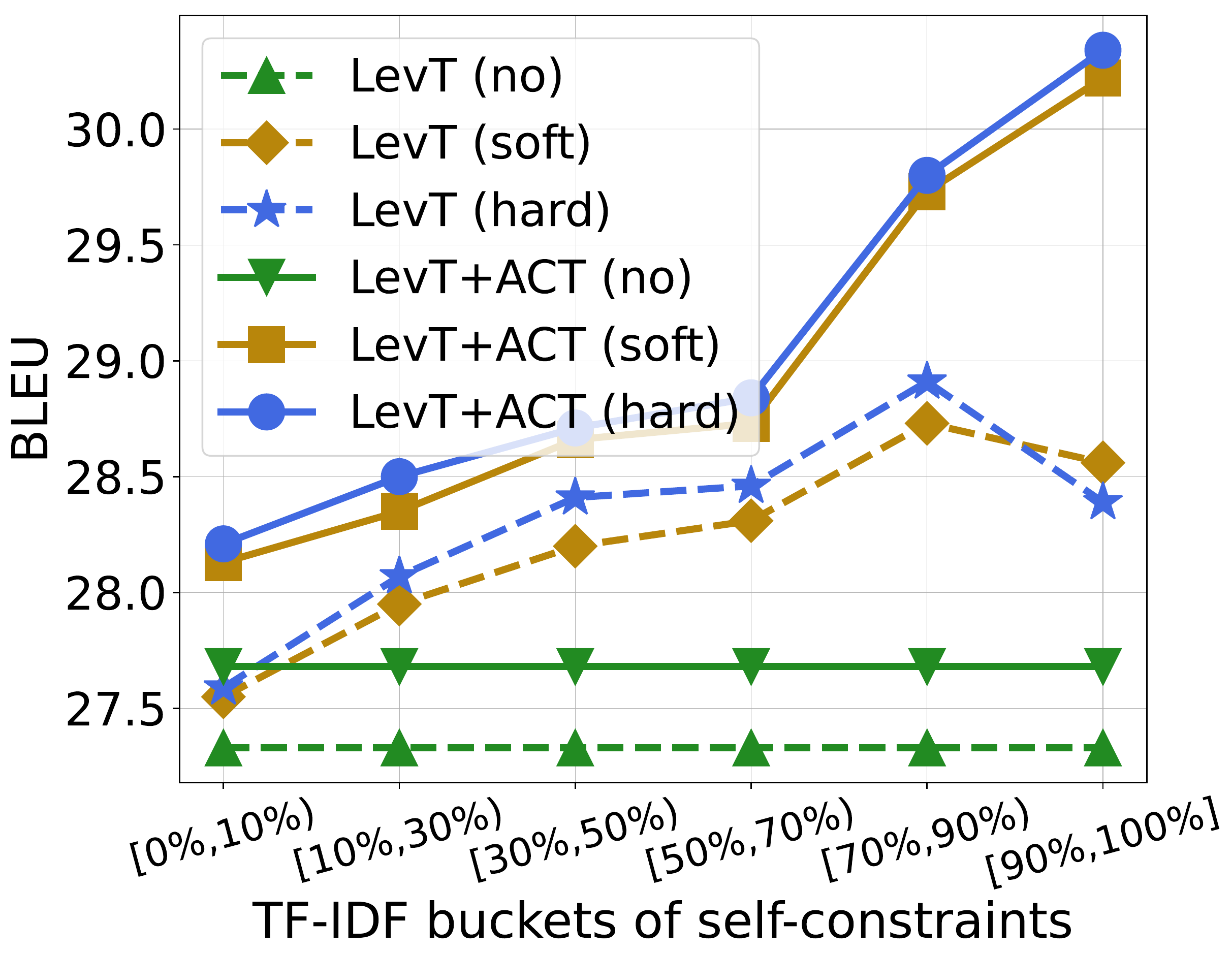}}
    \caption{Extended self-constrained translation results on WMT14-WIKT by removing UNK as constraints.
    The settings are the same as in $\mathsection$\ref{sec:self_constr_revisited}.
    }
    \label{fig:revision_self_constr}
\end{figure*}

\section{Unraveling the Buckets in Self-Constrained Translation}
\label{appendix:bucket}

In this section, we dig further into the buckets in self-constrained translation ($\mathsection$\ref{subsec: exploring_the_constraint_frequency}, $\mathsection$\ref{sec:self_constr_revisited}), especially for understanding why Drop\#1 happens.

As seen in Table \ref{tab:pos-result}, we categorize and count the constraints into five classes based on their Part-Of-Speech tagging with NLTK \cite{bird2009natural}.
We find that, 
1) punctuation (PUNK) dominates bucket 1;
2) as the constraint frequency decreases (from bucket 1 to bucket 6), the number of constraints identified as nouns (NN*) grows;
3) bucket 2 has the most UNK constraints.
The third finding is because, as the BPE training was only done on the training set of the datasets, there will be \texttt{<UNK>} on the target side of the test set.
Thus, cases in bucket 2 have a relatively large number of UNK tokens as constraints, resulting in the Drop\#1.

To give a clearer view about how is UNK causing Drop\#1, we exclude samples with UNK as constraints, and obtain a revised self-constrained translation results, as in Figure \ref{fig:revision_self_constr}.
Clearly, Drop\#1 disappears in the given setting.
Of course, Drop\#2 still verifies our claim in the paper.

% \begin{figure}
%    \centering
%    \includegraphics[width=0.9\linewidth]{figs/Revision Number of Constraints.pdf}
%    \caption{Corrected ablation results of constrained translation with one-to-multiple constraints.}
%    \label{fig:revision_self_constr_number}
%\end{figure}

\end{document}